\documentclass{amia}
\usepackage{lipsum} 
\usepackage{amsmath}
\usepackage{algpseudocode}
\usepackage{algorithm}
\usepackage{adjustbox}
\usepackage{color}
\usepackage{enumitem}
\usepackage{makecell}
\usepackage{hyperref} 
\usepackage{xcolor}
\usepackage[symbol]{footmisc}

\hypersetup{
    colorlinks,
    linkcolor={red!50!black},
    citecolor={blue!10!black},
    urlcolor={blue!50!black}
}

\setlist{topsep=0pt, leftmargin=*}

\begin{document}

\title{Knowledge-Driven Feature Selection and Engineering for Genotype Data with Large Language Models 
}

\author{
Joseph Lee, BS$^{1,*}$, Shu Yang, PhD$^{1,*,\dag}$, Jae Young Baik, Xiaoxi Liu, PhD$^{2}$, Zhen Tan, MS$^{3}$, Dawei Li, MS$^{3}$, Zixuan Wen, MA$^{1}$, Bojian Hou, PhD$^{1}$,
Duy Duong-Tran, PhD$^{4}$,\\ Tianlong Chen, PhD$^{5}$, Li Shen, PhD$^{1,\dag}$}

\def\thefootnote{${*}$}\footnotetext{Equal contribution by JL and SY. $^\dag$Correspondence: shu.yang@pennmedicine.upenn.edu (SY) and li.shen@pennmedicine.upenn.edu (LS).}

\institutes{
    $^1$ Unversity of Pennsylvania, Philadelphia, USA.
    $^2$  RIKEN, Yokohama, Japan.
    $^3$ Arizona State University, Tempe, USA.
    $^4$ United States Naval Academy, Annapolis, USA.
    $^5 $The University of North Carolina at Chapel Hill, Chapel Hill, USA
}

\maketitle

\section*{Abstract}
\textit{Predicting phenotypes with complex genetic bases based on a small, interpretable set of variant features remains a challenging task. 
Conventionally, data-driven approaches are used for this task, yet the high dimensional nature of genotype data makes the analysis and prediction difficult. 
Motivated by the biomedical knowledge encoded in pre-trained LLMs and the emerging applications for genetics, we set to examine the ability of LLMs in feature selection and engineering for tabular genotype data, with a novel knowledge-driven framework. 
We develop \textsc{FreeForm}, \textit{\underline{F}ree-flow \underline{R}easoning and \underline{E}nsembling for \underline{E}nhanced \underline{F}eature \underline{O}utput and \underline{R}obust \underline{M}odeling}, designed with chain-of-thought and ensembling principles, to select and engineer features with the intrinsic knowledge of LLMs.
Evaluated on two distinct genotype-phenotype datasets, genetic ancestry and hereditary hearing loss, we find this framework outperforms several data-driven methods, particularly on low-data regimes. 
\textsc{FreeForm} is available as open-source framework at GitHub: \url{https://github.com/PennShenLab/FREEFORM}.
}

\section*{Introduction}
\label{sec:methods}
Predicting observable phenotypes from genotype data has proven to be a monumental task in the field of genetics, with diverse applications ranging from personalized medicine \cite{torkmani2018} to genomic selection of crops\cite{guo2023machine}. Vast amounts of genetic variants, such as single nucleotide polymorphisms (SNPs) from high-throughput sequencing data, are often harnessed to make predictions on the phenotype. To deal with the sparsity of this data, linear models with regularization have been developed to much success in the case of polygenic risk scores \cite{torkmani2018, ma2021genetic}. Furthermore, machine learning models have been developed to capture the entangled epistatic relationships between genes and to predict complex traits \cite{medvedev2022human}. 

The difficulty, however, of modeling genotype data is substantial: 
(1) First, the topic of interest is not merely predictive (e.g. discovering causal variants\cite{uffelmann2021genome}) and machine learning methods are known to identify spurious features as significant due to multicollinearity \cite{krzywinski2015multiple}. Analyzing interaction terms---like epistatic relationships between variants---also introduces challenges such as a lack of interpretability when using complex models or multiple testing when examining higher-order interactions \cite{lippert2013exhaustive}. (2) Furthermore, a table of genotype data could contain thousands or even millions of columns. This `curse of dimensionality' can cause severe overfitting and amplify many issues, including those we've named above\cite{altman2018curse}. (3) Lastly, data can be limited in real settings, exacerbating existing concerns with overfitting.

Feature selection and feature engineering (i.e. feature construction) can be crucial steps to mitigate these concerns. Data-driven methods, such as Lasso regression, have demonstrated great success in selecting features \cite{pudjihartono2022review}. Feature engineering improves predictive performance without resorting to complex models and can help uncover interactions between features \cite{lou2013accurate}. However, these methods have their own issues: data-driven methods can struggle with small sample sizes and feature engineering is a laborious process that requires expertise to avoid multiple testing.

Recent advances in large language models (LLMs) have shown promise in addressing these challenges. With their remarkable performance across various tasks, LLMs have established themselves as powerful tools across many domains~\cite{achiam2023gpt,chang2023survey}. A key strength of LLMs lies in the knowledge they acquire through pre-training, enabling them to act as domain experts; recent LLMs have showed extensive understanding of biomedical concepts~\cite{singhal2023large, nori2023can, hou2024assessing}. Furthermore, LLMs can be greatly enhanced with well-designed prompting strategies~\cite{wei2022chain,wang2022self,yao2024tree}. Chain-of-thought prompting (CoT) ~\cite{wei2022chain} improves the reasoning of LLMs by encouraging step-by-step problem solving. Self-consistency \cite{wang2022self} addresses the naive greedy decoding used in CoT prompting by selecting the most consistent outcome across multiple reasoning paths.

Several studies have proposed to employ such capabilities of LLMs to perform feature preprocessing. Among them, Choi et al.~\cite{choi2022lmpriors} adopts the notion of prior knowledge in LLMs to conduct feature selection and casual discovery. Jeong et al.~\cite{jeong2024llm} examines three types of selection strategies with LLMs e.g. ranking vs. scoring features. Hollmann et al.~\cite{hollmann2024large} has an agentic approach, using the LLM to generate Python code that creates features in an iterative fashion based on cross-validation feedback. Han et al.~\cite{han2024large} employs LLMs to do feature engineering, generating conditional rules for each class label (e.g. age $>$ 21 increases the logits for label $0$) and repeating this several times to form an ensemble. 

Due to the promising performance of LLMs, there has also been a growing exploration of LLMs' usage in biomedical applications, especially genetics~\cite{toufiq2023harnessing, hou2024assessing, wang2024geneagent, shringarpure2024large}. Despite these innovations, most genetic studies have focused on gene-level data, likely due to the limitations of earlier LLMs. To our knowledge, no prior work has explored LLMs on variant-level data, except one that used API calls on NCBI databases to retrieve SNP information~\cite{jin2024genegpt}. While this approach compensates for LLMs’ inability to pass the GeneTuring benchmark~\cite{Hou2023.03.11.532238}, we argue the benchmark is an inadequate test of their utility, as it evaluates gene-SNP association by randomly sampling 100 SNPs from hundreds of millions. This setup overlooks how well LLMs leverage their knowledge of known variants (Fig. \ref{fig:qbox}), a gap we address in this study.

Inspired by these recent explorations~\cite{jeong2024llm, han2024large, toufiq2023harnessing,li2024exploring}, we propose to leverage LLMs' knowledge, intrinsic and augmented, and reasoning capabilities~\cite{wei2022chain,huang2022towards} to select the most informative genetic variants and generate novel features. 
We develop a knowledge-driven framework \textsc{FreeForm}, \textit{\underline{F}ree-flow \underline{R}easoning and \underline{E}nsembling for \underline{E}nhanced \underline{F}eature \underline{O}utput and \underline{R}obust \underline{M}odeling}, designed from ground principles in ensembling (self-consistency)\cite{wang2022self} and ``free-flow reasoning"\cite{tam2024let} to best leverage the expertise of LLMs. In the LLM-enabled framework, we implement scalable feature selection strategies that can process a large number of variants and feature engineering approaches that focus on interaction terms which are more interpretable.
We evaluate \textsc{FreeForm} on two real genotype datasets, genetic ancestry and hereditary hearing loss, and compare with data-driven and LLM-enabled methods. In particular, we focus on few-shot (\textit{i.e.} few data samples) settings where data-driven methods struggle due to limited sample size but LLMs have shown surprising generalizability~\cite{brown2020language,agrawal2022large}. In our study, we expect challenges when applying LLMs to tabular genotype data due to their limited semantics---column names are variant IDs and values indicate the number of minor allele copies ({0,1,2}). Thus, we analyze the effect of retrieval augmentation and domain-specific serialization when prompting. 

Our results highlight \textsc{FreeForm}'s potential to address three challenges present in modelling genotype data: Our method (1) enhances prediction while upholding interpretability, (2) reduces the dimensionality of the dataset, and (3) excels in low-shot regimes compared to data-driven methods by grounding ours in knowledge. Furthermore, we challenge the notion that LLMs lack knowledge of genetic variants by novelly applying them in this domain.

\section*{Methods}\label{sec:methods}

\begin{figure*}[h!]
\centerline{\includegraphics[width=1.05\textwidth]{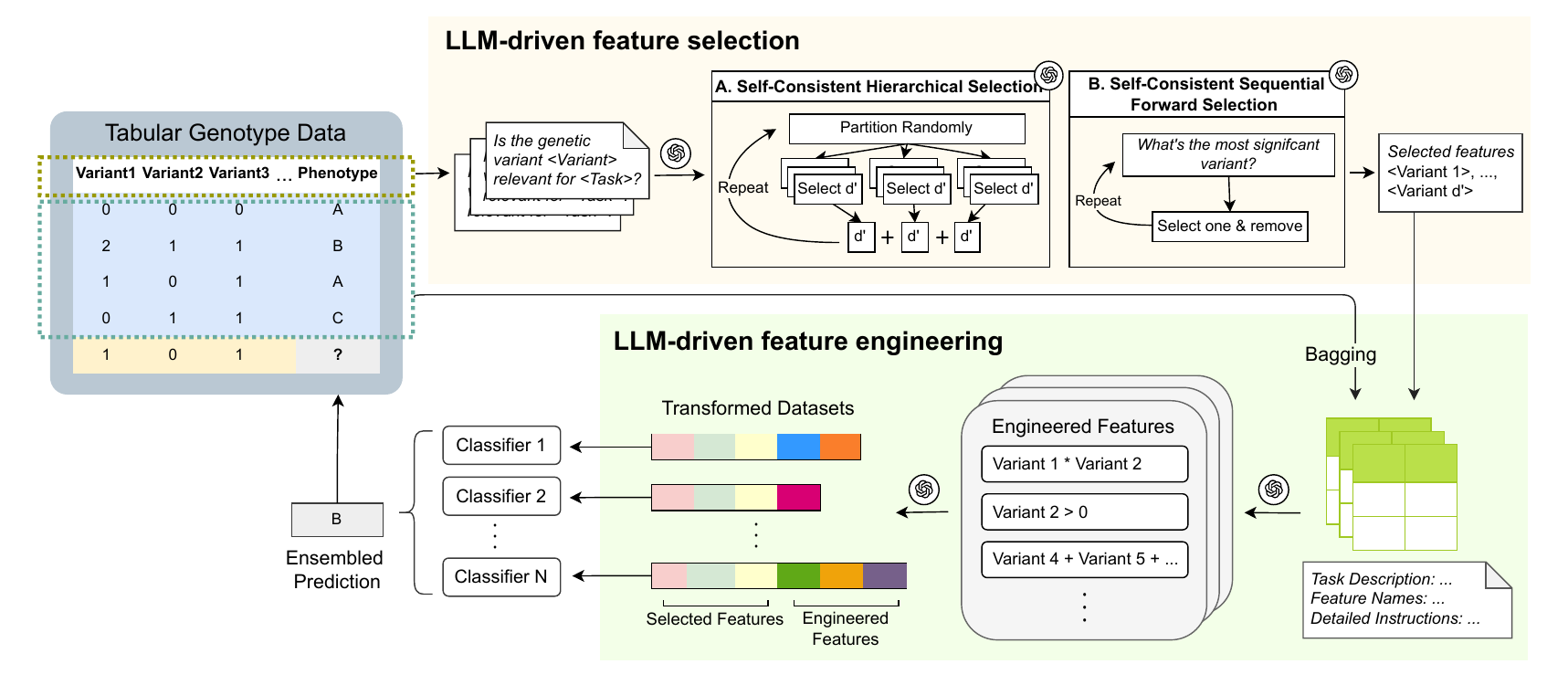}}
\vspace{-1.5mm}
\caption{Overview of the \textsc{FreeForm} framework. The pipeline consists of two parts: (1) LLM-driven feature selection takes $d$ variants and selects $d'$ of them (2) Given the selected features, we use LLMs to generate sets of engineered features to create an ensemble of classifiers.} 
\label{fig:fig1}
\end{figure*}

In this section, we introduce our \textsc{FreeForm} framework (Fig.~\ref{fig:fig1}), designed to address the challenges of training models on genotype data. The framework is built on two components: (1) leveraging the knowledge of LLMs to select a set of features and (2) leveraging the knowledge of LLMs to engineer new features from the selected features. 
\subsection*{Feature Selection}

The genotype dataset can be formalized as $D = \{(\mathbf{x}^i, y^i)\}^N_{i=1}$, i.e. a table comprising of $N$ labeled samples (\textit{i.e.} rows) each with $d$ variants (\textit{i.e.} columns). Each entry $x^i_j \in \{0,1,2\}$ represents the number of minor allele copies, while the corresponding label $y^i$ denotes the phenotype, e.g. $y^i \in \{$``African'', ``American'', ``East Asian'', ``European'', ``South Asian''$\}$. The column names, denoted by $S = \{s_j\}^d_{j=1}$, are text strings representing  variants such as rsIDs. 

The goal of feature selection is to identify a subset $S' \subseteq S$ with $d' \ll d$ variants such that a downstream model $f$, trained on $D_{S'} = \{(\mathbf{x}^i_{S'}, y^i)\}^N_{i=1}$, can make efficient, interpretable predictions. We use an LLM, modelled as a stochastic operator $\mathcal{T}$ parameterized by $\theta$ and given prompt $p$, to output a selected subset of features $S'$:
\begin{equation}
S' = \mathcal{T}(S,p; \theta),
\end{equation}
Notably, the proposed selection method is model agnostic unlike model-based rankings. Furthermore, previous LLM-based feature selection methods have not been tested in the setting of variants-level data where hallucination can be more likely~\cite{kandpal2023large} and high dimensionality is an issue. While context windows have become longer, new challenges have emerged \cite{liu2024lost, kortukov2024studying} in which naive usage of the context would be ineffective; to address these challenges, we design our feature selection strategies to scale for high dimensions while remaining token-efficient. In the downstream analysis, we train two models, Random Forest and Logistic Regression, on the selected features to evaluate their quality.

\paragraph{Relevance Filtering} We first ask the LLM to determine whether each of the $d$ variants is relevant to the task, requesting a simple “Yes” or “No”, yielding a subset of variants. This set may still be large, so we adjust the language of the prompt appropriately (e.g. ``potentially relevant") based on how many are filtered.

\paragraph{Self-Consistent Hierarchical Selection} is the first strategy we employ for selecting \(d'\) features from the filtered set of variants. We begin by randomly partitioning the variants into buckets of approximately 50 to 100 variants (a hyperparameter) to prevent the loss of information when contexts become too large\cite{liu2024lost}. Each bucket is independently passed to the LLM, tasked with selecting the $d'$ most relevant variants; we select $d'$ at every step in the case that the relevant features are concentrated in a single bucket. The $d'$ selected variants from each bucket are merged together, and the process repeats as delineated in Fig. \ref{fig:fig1}. We observe that the selection process is sensitive to the order in which features are presented to the LLM. Thus, for each bucket, we conduct multiple iterations, randomizing the order of variants and using a temperature of 0.3. This approach naturally integrates self-consistency \cite{wang2022self} by retaining the top \(d'\) variants that appear the most across iterations. During the final selection, we enhance the LLM with chain-of-thought (CoT) prompting, increase the temperature to 0.7, and increase the number of iterations to ten.

\paragraph{Self-Consistent Sequential Forward Selection} is the second strategy 
we employ. Starting with the filtered set of variants, we task the LLM with identifying the single most relevant variant with CoT. After each selection, the chosen variant is removed and the process repeats. Initially, we perform this extraction without any ensembling; the top few features are easy for the LLM to identify, but the task becomes more challenging as it becomes ambiguous which variants are more significant. After selecting a few, we start to apply self-consistency: repeating the extraction several times, in increasing amounts as we near the end of the selection. We find that this also mitigates the LLM’s tendency to return the features that were already selected. In cases where the LLM still fails to identify a valid variant, which occurs frequently towards the end, we implement an exception handling mechanism that retries the selection process.

\subsection*{Feature Engineering}

Given the selected features $S'$, our goal is to engineer new features that capture meaningful relationships the model might not identify on its own (e.g.,  $\text{household density} = \text{family size} \div \text{number of rooms}$). We automate this traditionally manual, expert-driven process by leveraging the knowledge and reasoning capabilities of LLMs. As outlined in Fig. \ref{fig:fig1}, we repeat this several times and train a model on each feature set, forming an ensemble.

Formally, we transform our dataset $D_{S'}$ into $K$ transformed datasets $D_{k} = \{(\mathbf{x}^i_{S'_k}, y^i)\}^N_{i=1}$, where $S'_k \supseteq S'$. Each transformed dataset is created by prompting the LLM $\mathcal{T}$ with $p'$, which includes a serialized representation of selected examples $R = \{(\mathbf{x}^{i}_{S'}, y^{i})\}_{i \in R \subseteq D_{S'}}$, to enhance the LLM with context \cite{brown2020language}. We define this function $\mathtt{Serialize}$ to convert each row to a textual description (e.g. ``The $s_1$ variant of the person has $x^i_1$ minor alleles$...$ The $s_{d'}$ variant of the person has $x^i_{d'}$ minor alleles.") as LLMs usually prefer natural language. 
Since LLMs can be sensitive to the input, we examine various serialization templates and prompts. To address gaps in the LLM's prior knowledge, we also explore retrieval-augmentation to 
supplement the variant IDs with its associated gene~\cite{lewis2020retrieval}.

The LLM $\mathcal{T}$ outputs the new features that are added back onto $S'$ to create $S'_k$:
\begin{equation}
S'_k = S' \cup \mathcal{T}\left(S', p'; \theta\right), \text{where } p' \text{ includes } \{\mathtt{Serialize}(\mathbf{x}^{i}_{S'}, y^{i}, S')\}_{i \in R}
\end{equation}
We then train $K$ models, $\mathcal{F} = \{f_k\}^K_{k=1}$, each on a dataset $D_k$, aiming to capture different hypotheses about how the variants relate to the phenotype. During inference, for input $\mathbf{x}$, we average the class probabilities $\mathbf{p}(f_k(\mathbf{x}))$ from each model $f_k$. The final prediction $\hat{y}$ is made by selecting the class $i$ with the highest averaged probability:
\vspace{-2mm}
\begin{equation}
\begin{aligned}
\mathbf{p}^*(\mathbf{x}) &= \frac{1}{K} \sum_{k=1}^{K} \mathbf{p}(f_k(\mathbf{x})),\\
\hat{y} &= \arg\max_i \, p_i^*(\mathbf{x}).
\end{aligned}
\vspace{-2mm}
\end{equation}
Our aim is to harness the diverse feature representations generated by the LLM, reducing the risk of overfitting in low-shot settings by anchoring each feature set in the knowledge embedded within  LLMs rather than the limited data. To uphold interpretability, we limit our feature construction to interaction terms. Notably, our method is model-agnostic, de-coupled from the classifier. We train two models, Random Forest and Logistic Regression, on the transformed datasets to evaluate the quality of the constructed features. We discuss key steps of our method in depth below.

\paragraph{Automating Feature Engineering}

\begin{figure}[h!]
\vspace{-3mm}
\centerline{\includegraphics[width=\textwidth]{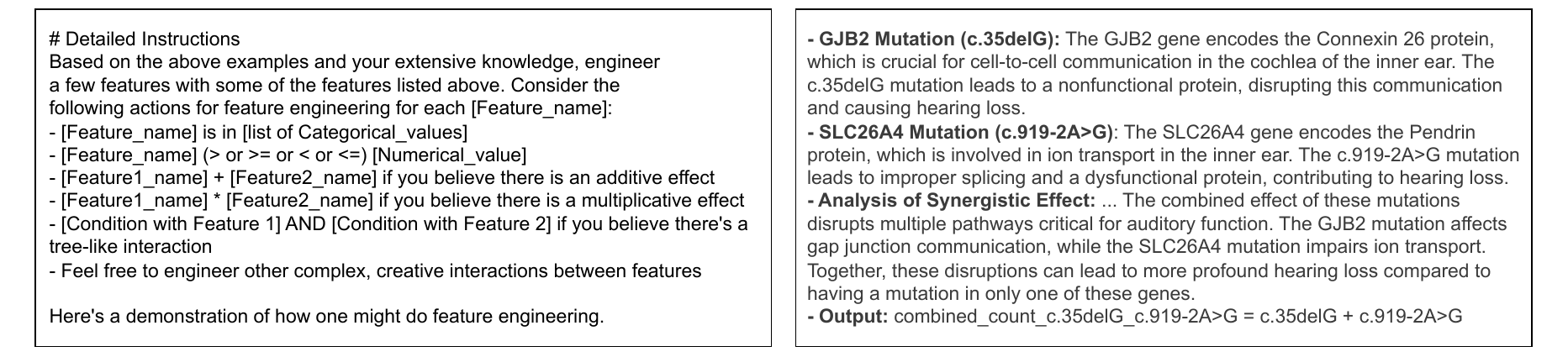}}
\vspace{-2mm}
\caption{Example of Detailed Instructions }

\label{fig2}
\end{figure}

When asking the LLM to engineer features, we provide a comprehensive prompt $p'$ that includes the following components:
\begin{itemize}[noitemsep,topsep=0pt,parsep=0pt,partopsep=0pt]
    \item Instructions: Directions to use the provided features to engineer new features relevant to the task.
    \item Task Description: A concise description of the specific task for which the features are being engineered.
    \item Features: A list of features including the name of each genetic variant.
    \item Examples: $|R|$ examples that illustrate the data in a serialized format.
    \item Detailed Instruction: List of specific choices for feature engineering, such as multiplying or adding features, accompanied by a task-specific demonstration as seen in Fig \ref{fig2}.
    \item Step-by-Step Solution: Directions for the LLM to solve the problem step by step.
\end{itemize}
\paragraph{Free-Flow Reasoning}
Traditionally, we enforce the output of the LLM to be structured e.g. JSON. However, we deliberately allow the LLM to freely generate its response, recognizing that enforcing a rigid structure can diminish the depth and quality of the LLM's CoT reasoning \cite{tam2024let}. To guide the LLM, we provide an example of a ``correctly'' engineered feature \cite{tong2024can}, which we self-generate with an LLM and then manually verify.

\paragraph{Self-Parsing and Function Writing via LLMs} The unstructured output generated by the LLM can present challenges for parsing. To address this, we employ the LLM itself to extract the engineered features from its output, listing them line by line for easy parsing. Subsequently, we task the LLM with writing an executable Python function that will generate the new columns of dataset $D_k$ based on the extracted features. When errors are detected, we implement error-handling mechanisms to catch any issues and they are fed back into the LLM which rewrites the function.

\paragraph{Ensembling}
To mitigate overfitting, we repeat the feature engineering process \( K \) times to generate all $D_k$ which we'll use to train $K$ classifiers. Similar to \textsc{FeatLLM} \cite{han2024large}, we incorporate additional ensembling and order bias mitigation strategies such as bagging and order shuffling, where we pass a random subset of $|R| \leq N$ samples to the LLM; we limit the number of samples $|R|$ to $ 16$. We do this to further diversify the LLM output for each iteration, and simultaneously avoid exorbitant usage of the context window. Our free-flow reasoning approach further contributes to the diversity of the output, as we naturally allow the LLM to determine the number of features that will be constructed during its generation. By setting a non-zero temperature of 1, we ensure that each iteration further produces a varied set of features, especially in type. The resulting \( k \) transformed datasets form the basis for the ensembled model.

Due to the page limit, we direct readers to our \href{https://github.com/PennShenLab/FREEFORM}{GitHub repo}, which contains the detailed prompts and relevant hyperparameters used in the entire pipeline. The source code of \textsc{FreeForm} is provided to promote reproducibility.

\section*{Results}

\label{sec:results}

\subsection*{Experiment Setup}

\paragraph{Datasets}
Our experiments involve two datasets: the \textbf{Genomic Ancestry Dataset}\cite{fairley2020international} and the \textbf{Hereditary Hearing Loss Dataset}\cite{luo2021machine}. The Genomic Ancestry Dataset is derived from the 1000 Genomes Project (1KGP). 
We focus on determining the superpopulation ancestry phenotype (African, American, East Asian, European, and South Asian). In particular, we used a curated set of 10,000 SNPs predefined by GRAF to pinpoint ancestry markers~\cite{Moustafa2023}, and we'll discuss how we addressed the issues that arise from their quality control (QC).  After QC and preprocessing, the dataset includes 2,403 subjects and  8,688 variants as columns in the rsID format, a standard identifier used by dbSNP. 

The Hereditary Hearing Loss Dataset is considerably smaller, comprising of 1,209 subjects and 144 variants as columns, employing the HGVS nomenclature system. This dataset is notably imbalanced, with approximately 75.9\% of the samples classified as ``Yes'' (indicating the presence of hereditary hearing loss) and 24.1\% as ``No.'' To our best knowledge, these two are the only open-access genotype datasets available online.

\paragraph{Baselines} 
For feature selection, we conduct comparisons with four baseline methods. The first three baselines are conventional machine learning approaches: (1) \textbf{LASSO}, (2) \textbf{PCA} and (3) \textbf{RF-based Gini Importance} where we fit a Random Forest on the training data and rank the features by their Gini importance. In our case, the specification of a fixed number of features is required for our study. LASSO, however, reduces an arbitrary number of coefficients to zero during model training. Similarly, PCA provides loadings along principal axes and has been widely used in the genetic field to select variants. To adapt these methods to our requirements, we select the $d'$ features with the largest coefficients in the highest-performing LASSO model. For PCA, we take the top $d'$ loadings of the first principle component. Lastly, we include (4) \textsc{LLM-Select}\cite{jeong2024llm} (we use their \textsc{LLM-RANK} prompts).

For feature engineering, we compared our approach against five baselines. The first three are traditional machine learning methods: (1) \textbf{Logistic Regression}, (2) \textbf{Random Forest} and (3) \textbf{XGBoost}. We also include recent baselines, (4) \textbf{TabPFN} \cite{hollmann2022tabpfn}, a foundation model for tabular data, and (5) \textbf{FeatLLM}\cite{han2024large} which also leverages LLMs to do feature engineering but limits themselves to conditional rules (e.g. variant 1 is $>$ 0) for a linear model.
\paragraph{Implementation Details}

Our \textsc{FreeForm} framework utilizes \texttt{GPT-4o} (\texttt{2024-05-13}) as the primary LLM backbone, particularly for tasks requiring advanced reasoning capabilities, such as automating feature engineering and selecting relevant features. For more routine tasks, including parsing output and writing Python functions, we employ \texttt{GPT-3.5-turbo} which offers a cost-effective solution that meets the performance requirements for these functions. These models are called upon using the OpenAI API  which only requires internet access.

For feature engineering, we employ an ensemble of $K=20$ models, striking a balance between cost-effectiveness and model performance, noting that performance gains diminish beyond this point. In replicating baselines, machine learning models were implemented using Python’s \texttt{scikit-learn} library. Hyperparameters were optimized using grid search and $k$-fold cross-validation, with \(k\) set to either 2 or 4, ensuring that the training set includes at least one example of each class. For other methods, such as \textsc{FeatLLM} and \textsc{TabPFN}, we used default parameters with slight adjustments for fair comparison (e.g., using 15 conditions instead of 10 for \textsc{FeatLLM}). Also, for the evaluation of all feature engineering methods, we choose one of the feature sets generated by hierarchical selection.

\subsection*{Main Results}

\begin{figure}[h!]
    \centering
    \includegraphics[width=0.8\textwidth]{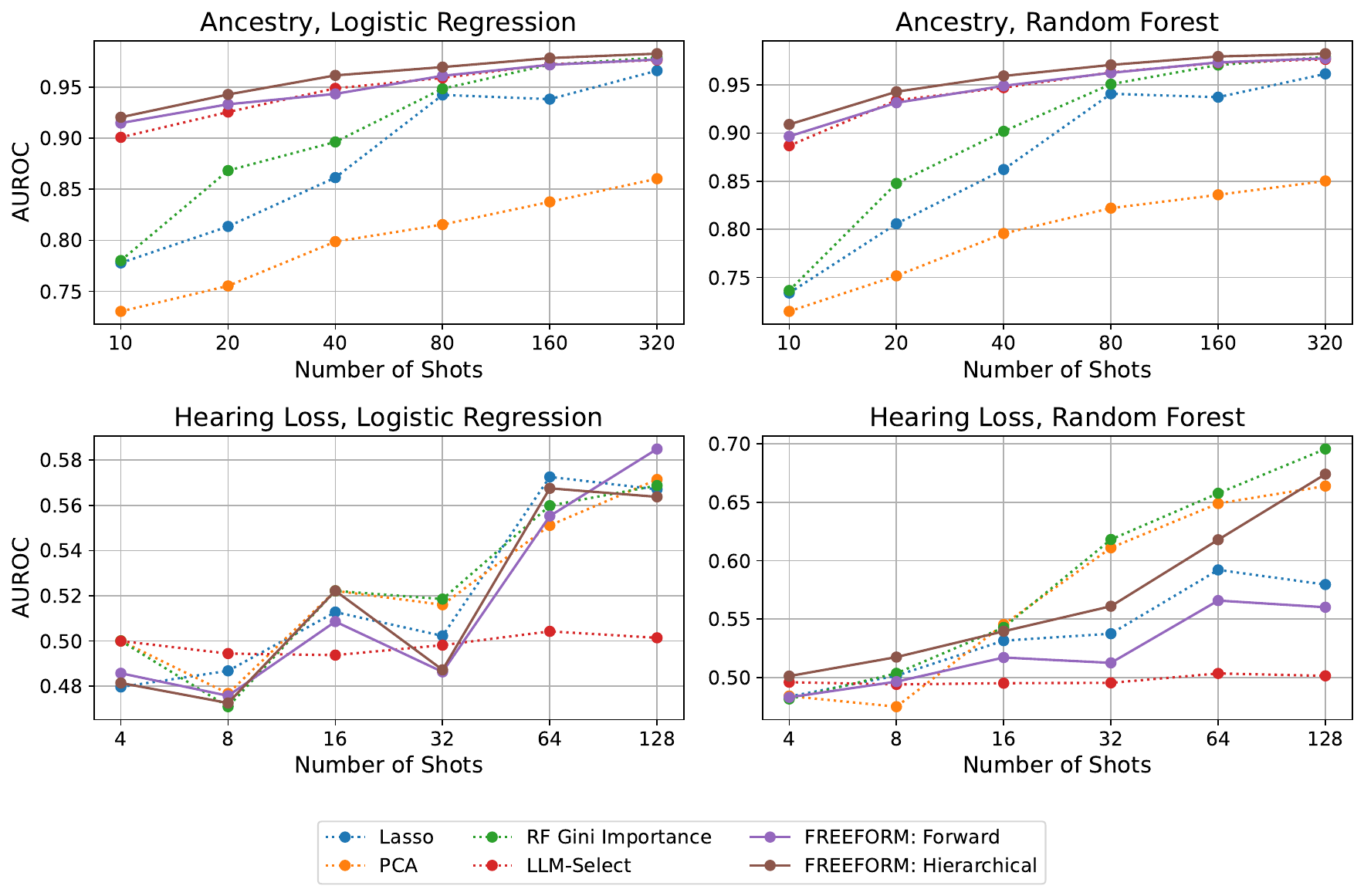}
 
    \caption{Evaluation of Feature Selection on Ancestry and Hearing Loss}
    
    \label{fig:fig3}
\end{figure}

In Fig~\ref{fig:fig3}, we compare the performance of our feature selection methods against baselines using Logistic Regression and Random Forest as downstream classifiers. For feature selection, we repeat our experiments five times with cross-validation, limiting the evaluation to few-shot settings
where $N \leq 320$ for ancestry and $N \leq 128$ for hearing loss. For data-driven methods, we use the training data (varying the size of $N$) to perform feature selection and train the classifier on the same training data. We also emphasize that, unlike the data-driven methods, the LLM-driven methods perform feature selection without relying on any data samples, leveraging only the model's prior knowledge. Their usage of the training data is limited to training the downstream classifier. 

Our findings indicate that LLM-driven methods significantly outperform data-driven approaches for feature selection in low-shot regimes, achieving gains of up to approximately 20\%. Notably, in the genomic ancestry task, LASSO requires 80 shots to achieve similar results to what our framework achieves with just 10 shots. In hearing loss, we observe that the performance gap is smaller and our advantage remains until 16 shots, when using Random Forests. This discrepancy is likely due to the limited presence of variants within the dataset that the LLM has knowledge of.

\begin{figure}[h!]
\vspace{-3mm}
    \centering
    \includegraphics[width=0.8\textwidth]{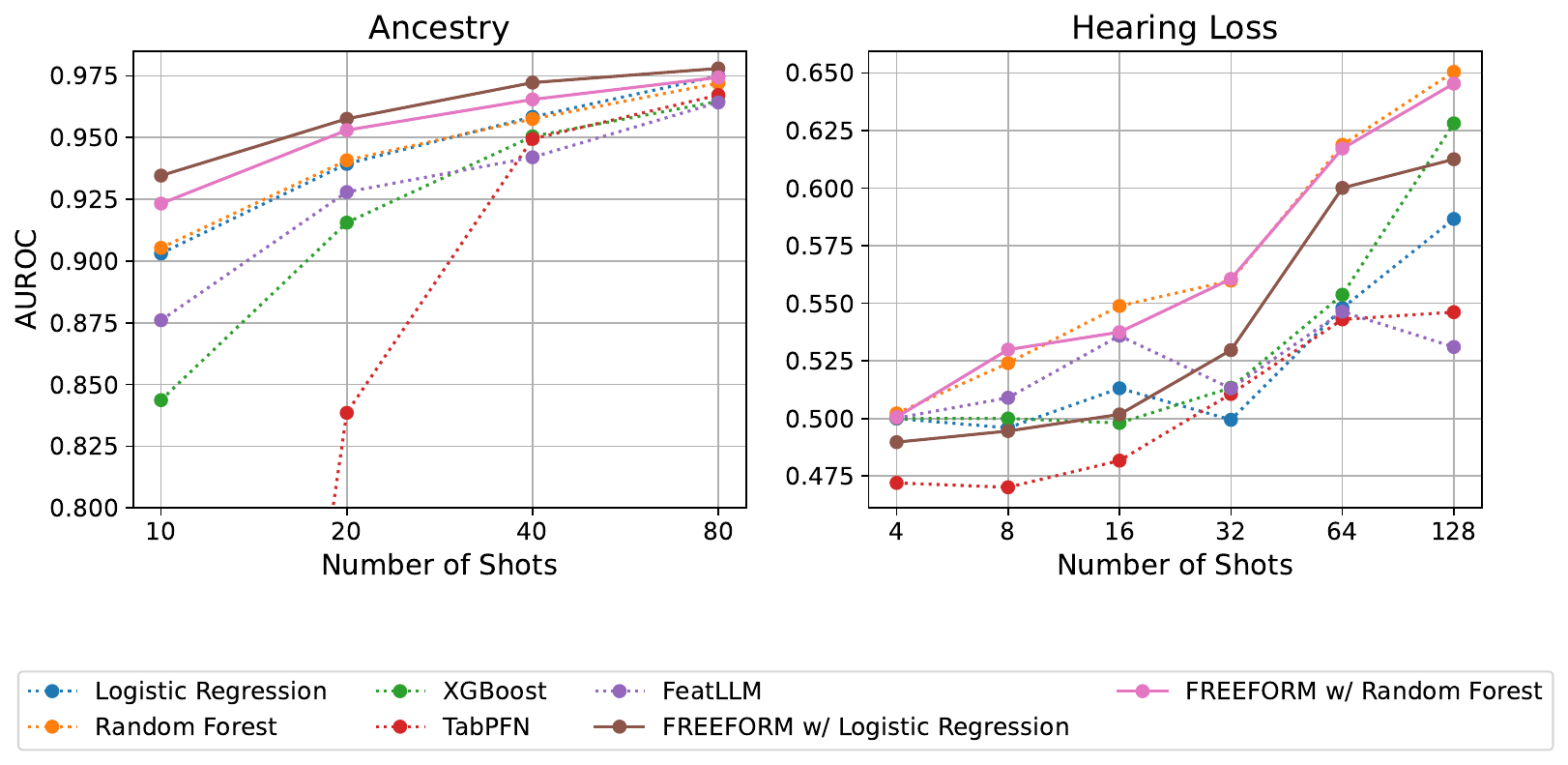}
   
    \caption{Evaluation of Feature Engineering on Ancestry and Hearing Loss}
    \label{fig:fig4}
\vspace{-2mm}
\end{figure}

For feature engineering, we conduct our experiments five times with cross-validation, limiting the evaluation to few-shot settings
where $N < 120$. In Fig~\ref{fig:fig4}, \textsc{FreeForm} consistently consistently ranks at or above baseline models. In the genomic ancestry task, our framework improves the performance of both Logistic Regression and Random Forest, especially in the lower-shot scenarios, and outperforms recent models such as \textsc{FeatLLM} and \textsc{TabPFN}. However, as the number of shots approaches 80, the gap between our methods and the baselines decreases. 
In the hearing loss task, our framework notably enhances the performance of Logistic Regression on higher-shot scenarios. While Random Forest does not benefit from the engineered features, 
our approach remains competitive. This outcome suggests limited effectiveness of the interaction terms we engineer for complex models like Random Forests. 

\subsection*{Ablations and Analysis}

\begin{figure}[h!]
\vspace{-4mm}
    \centerline{\includegraphics[width=\textwidth]{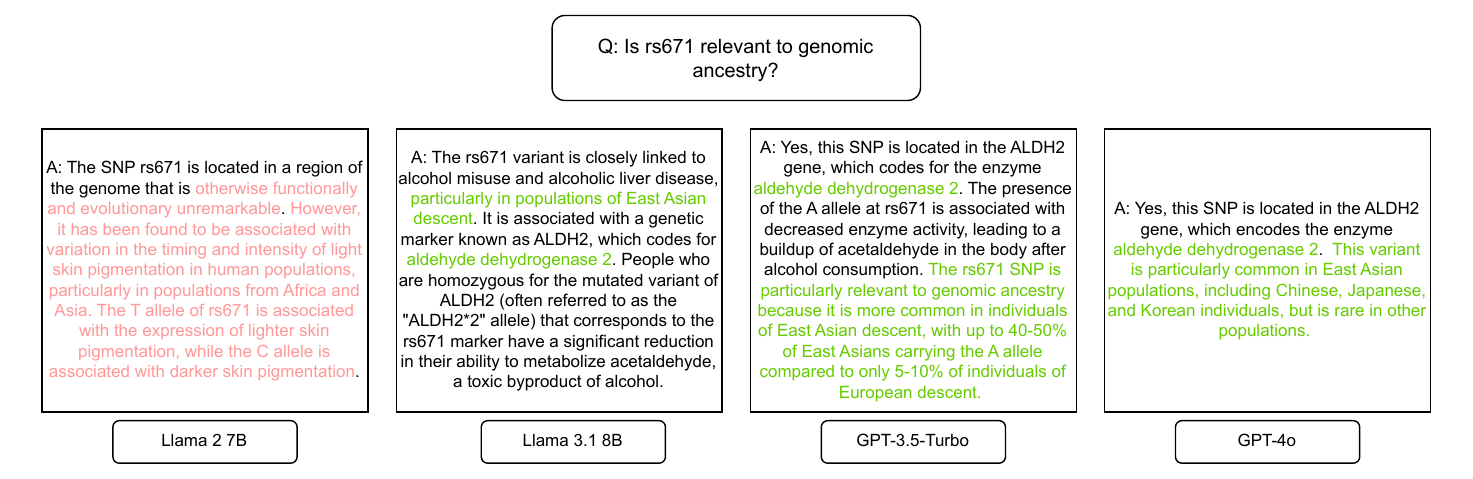}}
\caption{Comparison between different LLM models on their knowledge of the SNP rs671 relating to genomic ancestry. Red text indicates a hallucination, which was only observed in the case of the Llama 2 7B model. }
\label{fig:qbox}
\end{figure}

\paragraph{Open Source Models}
In Table \ref{table:opensource}, we examine the performance of various LLMs in our framework, repeating the experiment five times with cross-validation. For feature selection, \texttt{gpt-4o} performs better as expected, except for hearing loss, further suggesting limited knowledge of the relevant variants across the LLMs. It remains unclear whether this advantage arises from the LLMs’ ability to identify relevant features or their depth of knowledge (e.g. while an LLM may recognize a variant, its expertise regarding that variant may vary). For feature engineering, the weaker models are highly competitive. This is surprising; this may indicate that the interaction terms we enable (e.g., conditional rules, multiplicative expressions) are within the reasoning capabilities of all studied models. Alternatively, it could reflect all the models’ inability to capture epistasis. Either way, these results align with the shared awareness of variants among models (see Figure \ref{fig:qbox}), suggesting that even basic knowledge can enable LLMs to act as weak learners. 

\begin{table}[h!]
\vspace{2mm}
\small
\centering
\caption{\textsc{FreeForm} using different models. For feature selection, hierarchical selection is used across all models. We limit the analysis to 16-shots (Hearing Loss) and 20-shots (Ancestry). LR: Logistic Regression. RF: Random Forest. Values shown are AUROC (AUC). Standard deviation is in parenthesis.}

{
\setlength{\tabcolsep}{3pt} 
\begin{tabular}{l|cc|cc|cc|cc}
\hline
\textbf{Model} & \multicolumn{4}{c|}{\textbf{Feature Selection}} & \multicolumn{4}{c}{\textbf{Feature Engineering}} \\ \hline
& \multicolumn{2}{c|}{\textbf{Hearing}} & \multicolumn{2}{c|}{\textbf{Ancestry}} & \multicolumn{2}{c|}{\textbf{Hearing}} & \multicolumn{2}{c}{\textbf{Ancestry}} \\  
 & \textbf{LR} & \textbf{RF} & \textbf{LR} & \textbf{RF} & \textbf{LR} & \textbf{RF} & \textbf{LR} & \textbf{RF} \\ \hline
GPT-3.5-turbo & \textbf{0.524} (0.03) & 0.539 (0.04) & 0.785 (0.03) & 0.765 (0.04) & 0.506 (0.10) & 0.540 (0.11) & 0.955 (0.01) & 0.951 (0.01) \\ 
Llama-3.1-405B &  --- &  --- &  --- &  --- & 0.511 (0.11) & \textbf{0.544} (0.11) & 0.956 (0.02) & 0.952 (0.01) \\ 
GPT-4o &  0.506 (0.09) &   \textbf{0.544} (0.10) &  \textbf{0.943} (0.01) &  \textbf{0.943} (0.02) & \textbf{0.514} (0.11) & 0.543 (0.11) & \textbf{0.957} (0.01) & \textbf{0.953} (0.01) \\ \hline
\end{tabular}
}
\vspace{-0mm}
\label{table:opensource}
\end{table}

\paragraph{Augmentation and Serialization}

 In our main results of \textsc{FreeForm}, we use a simple serialization strategy like the following:  ``$s_1$ is $x^i_1$. $s_2$ is $x^i_2$... Answer: $y_i$''. In this ablation study, we find that using a more elaborate schema like ``The $s_1$ variant of the person has $x^i_1$ minor alleles $...$" does not make a difference. While this may be surprising given the existing efforts on exploring serialization strategies \cite{hegselmann2023tabllm}, large foundation models may be robust to such formatting. Furthermore, our findings show that the augmentation of gene information is not significant. Our experimentation is limited to providing the genes associated with the variant. This may be redundant information for the LLM but other strategies are not straightforward; augmenting literature for a variant is challenging due to the lack of relevance it may have to the task. This will be an important avenue to explore for future work.
\begin{table}[h!]
\small
\centering
\caption{\textsc{FreeForm} using genotype-specific strategies for featuring engineering on Hearing Loss (16-shot) and Ancestry (20-shot). LR: Logistic Regression. RF: Random Forest.}
\begin{tabular}{lcc}
\hline
\textbf{Configuration} & \textbf{LR AUC (Std)} & \textbf{RF AUC (Std)} \\ \hline
\textbf{Hearing Loss} & & \\ \hline
FreeForm: Feature Engineering & \textbf{0.5145} (0.11) & \textbf{0.5438} (0.11) \\
+ Genotype Serialization & 0.5127 (0.11) & 0.5384 (0.11) \\
+ Gene Augmentation & 0.5110 (0.09) & 0.5387 (0.11) \\ \hline
\textbf{Ancestry} & & \\ \hline
FreeForm: Feature Engineering & \textbf{0.9572} (0.01) & 0.9527 (0.01) \\
+ Genotype Serialization & 0.9568 (0.01) & \textbf{0.9532} (0.01) \\
+ Gene Augmentation & 0.9571 (0.01) & 0.9530 (0.01) \\ \hline
\end{tabular}
\vspace{-4mm}
\label{table:freeform_results}
\end{table}
\paragraph{Feature Nomination}
The genomic ancestry dataset we used is a curated version of the full dataset, after applying rigorous quality control. While this ensures data reliability, it can lead to the omission of significant genetic variants. To address this concern, we ask \texttt{GPT-4o} to suggest fifteen SNPs; SNPs such as rs671 or rs2814778 with clear causal relationships or statuses as standard AIMs were usually suggested. These SNPs originally existed in the database but were omitted in the curation of the 10K version so we inserted them back. Our LLM-driven selection methods were able to recover most of these variants while the data-driven methods could not, and we found that these variants contributed largely to the performance gap we observed for feature selection in the genomic ancestry dataset. We find it concerning that data-driven methods failed to identify many of these variants despite their predictive power. In additional analysis, training a Logistic Regression model with the \texttt{GPT-4o}-suggested SNPs alone achieved an average of 0.94 AUC on 20 shots whereas a set of fifteen SNPs selected by PCA on the original dataset achieved 0.78 AUC. LLMs could be a promising method to mitigate the issues of quality control, a frequent issue in this domain, providing a potentially more robust, automatable method than manually imputing well-known variants from the literature. 

\section*{Discussion}

We present \textsc{FreeForm} which advances the state-of-the-art in LLM-based, few-shot tabular learning and we novelly apply our LLM-driven framework to genotype data. \textsc{FreeForm} goes beyond the typical usage of LLMs for inference and aids the process of feature selection and engineering, tackling the issues of high dimensionality, limited data samples, and interpretability. Furthermore, we find that LLMs have a robust knowledge of genetic variants, demonstrating state-of-the-art performance across different variant ID schemas, showcasing the promise of LLMs in genetics.

Our framework notably has several key advantages over existing approaches besides performance: (1) it is model-agnostic, (2) it scales well to higher dimensions, and (2) it incurs no inference costs, as features are engineered once during training, unlike LLM-only methods with high computational costs at inference time. Moreover, the entire pipeline can be executed using API access, costing approximately one dollar to run, with the majority of the pipeline completing within minutes, aside from the initial filtering step in feature selection. 

However, \textsc{FreeForm} has room for growth. The current results suggest we either allow a naive range of variants interaction types in feature engineering, or none of the models are capable of capturing epistasis. Our retrieval augmentation is also limited to gene information. Additionally, the LLM’s input could be enhanced by including high-level feature statistics typically considered in this domain. Future work remains in expanding its capabilities to generate novel features, improve the extraction of knowledge intrinsic to LLMs, better augment task-specific knowledge from APIs such as PubMed, and integrate further interpretability or explainability. Recent studies have demonstrated the potential for LLMs in the discovery of new gene sets or causal genes \cite{wang2024geneagent, shringarpure2024large}, where LLMs effectively interpolate across the vast corpus of scientific literature they are trained on. We find this approach promising for addressing the issue of multicollinearity, which we acknowledged but did not resolve in this work. Furthermore, we see promising future work in developing feature nomination, where we used the LLM to suggest predictive features. The controllability of LLMs through prompting opens up interesting possibilities, such as the nomination of features that better represent diverse populations, thereby mitigating biases that data-driven methods can exacerbate. One limitation of our study is that a few assumptions were made in the evaluation, including the fixed selection of fifteen features, to showcase our pipeline. Thus, we plan to expand our evaluation to more scenarios and more phenotypes, such as Alzheimer's disease, to demonstrate its robust utility. 

As LLMs advance in domain expertise, potentially surpassing humans \cite{luo2024large}, their potential to revolutionize bioinformatics becomes increasingly imminent. While our study demonstrates LLMs' excellence in low-shot regimes, we acknowledge such scenarios are rare in practice. We anticipate, however, these capabilities will scale as foundational models advance and domain-specific LLMs develop. Recent efforts, such as fine-tuning foundational models on literature \cite{luo2024large} or augmenting LLMs with knowledge graphs \cite{li2024dalk}, are making progress towards this. While our framework focuses on feature selection and engineering, our work serves as a prototype, showcasing the potential of LLMs in genetics.

\subparagraph{Acknowledgments} This work was supported in part by the NIH grants U01 AG066833, U01 AG068057, R01 AG071470, U19 AG074879, and S10 OD023495. 
\makeatletter
\renewcommand{\@biblabel}[1]{\hfill #1.}
\makeatother
\small
\bibliographystyle{vancouver}
\bibliography{amia}  

\begin{thebibliography}{10}

\bibitem{torkmani2018}
Torkmani A, Wineinger NE, Topol EJ.
\newblock The personal and clinical utility of polygenic risk scores.
\newblock Nature Reviews Genetics. 2018;19:581-90.

\bibitem{guo2023machine}
Guo T, Li X.
\newblock Machine learning for predicting phenotype from genotype and environment.
\newblock Current Opinion in Biotechnology. 2023;79:102853.

\bibitem{ma2021genetic}
Ma Y, Zhou X.
\newblock Genetic prediction of complex traits with polygenic scores: a statistical review.
\newblock Trends in Genetics. 2021;37(11):995-1011.

\bibitem{medvedev2022human}
Medvedev A, Mishra~Sharma S, Tsatsorin E, Nabieva E, Yarotsky D.
\newblock Human genotype-to-phenotype predictions: Boosting accuracy with nonlinear models.
\newblock PloS one. 2022;17(8):e0273293.

\bibitem{uffelmann2021genome}
Uffelmann E, Huang QQ, Munung NS, De~Vries J, Okada Y, Martin AR, et~al.
\newblock Genome-wide association studies.
\newblock Nature Reviews Methods Primers. 2021;1(1):59.

\bibitem{krzywinski2015multiple}
Krzywinski M, Altman N.
\newblock Multiple linear regression: when multiple variables are associated with a response, the interpretation of a prediction equation is seldom simple.
\newblock Nature methods. 2015;12(12):1103-5.

\bibitem{lippert2013exhaustive}
Lippert C, Listgarten J, Davidson RI, Baxter J, Poon H, Kadie CM, et~al.
\newblock An exhaustive epistatic SNP association analysis on expanded Wellcome Trust data.
\newblock Scientific reports. 2013;3(1):1099.

\bibitem{altman2018curse}
Altman N, Krzywinski M.
\newblock The curse (s) of dimensionality.
\newblock Nat Methods. 2018;15(6):399-400.

\bibitem{pudjihartono2022review}
Pudjihartono N, Fadason T, Kempa-Liehr AW, O'Sullivan JM.
\newblock A review of feature selection methods for machine learning-based disease risk prediction.
\newblock Frontiers in Bioinformatics. 2022;2:927312.

\bibitem{lou2013accurate}
Lou Y, Caruana R, Gehrke J, Hooker G.
\newblock Accurate intelligible models with pairwise interactions.
\newblock In: Proceedings of the 19th ACM SIGKDD international conference on Knowledge discovery and data mining; 2013. p. 623-31.

\bibitem{achiam2023gpt}
Achiam J, Adler S, Agarwal S, Ahmad L, Akkaya I, Aleman FL, et~al.
\newblock Gpt-4 technical report.
\newblock arXiv preprint arXiv:230308774. 2023.

\bibitem{chang2023survey}
Chang Y, Wang X, Wang J, Wu Y, Yang L, Zhu K, et~al.
\newblock A survey on evaluation of large language models.
\newblock ACM Transactions on Intelligent Systems and Technology. 2023.

\bibitem{singhal2023large}
Singhal K, Azizi S, Tu T, Mahdavi SS, Wei J, Chung HW, et~al.
\newblock Large language models encode clinical knowledge.
\newblock Nature. 2023;620(7972):172-80.

\bibitem{nori2023can}
Nori H, Lee YT, Zhang S, Carignan D, Edgar R, Fusi N, et~al.
\newblock Can generalist foundation models outcompete special-purpose tuning? case study in medicine.
\newblock arXiv preprint arXiv:231116452. 2023.

\bibitem{hou2024assessing}
Hou W, Ji Z.
\newblock Assessing GPT-4 for cell type annotation in single-cell RNA-seq analysis.
\newblock Nature Methods. 2024:1-4.

\bibitem{wei2022chain}
Wei J, Wang X, Schuurmans D, Bosma M, Xia F, Chi E, et~al.
\newblock Chain-of-thought prompting elicits reasoning in large language models.
\newblock Advances in neural information processing systems. 2022;35:24824-37.

\bibitem{wang2022self}
Wang X, Wei J, Schuurmans D, Le Q, Chi E, Narang S, et~al.
\newblock Self-consistency improves chain of thought reasoning in language models.
\newblock arXiv preprint arXiv:220311171. 2022.

\bibitem{yao2024tree}
Yao S, Yu D, Zhao J, Shafran I, Griffiths T, Cao Y, et~al.
\newblock Tree of thoughts: Deliberate problem solving with large language models.
\newblock Advances in Neural Information Processing Systems. 2024;36.

\bibitem{choi2022lmpriors}
Choi K, Cundy C, Srivastava S, Ermon S.
\newblock LMPriors: Pre-Trained Language Models as Task-Specific Priors.
\newblock In: NeurIPS 2022 Foundation Models for Decision Making Workshop;. .

\bibitem{jeong2024llm}
Jeong DP, Lipton ZC, Ravikumar P.
\newblock LLM-Select: Feature Selection with Large Language Models.
\newblock arXiv preprint arXiv:240702694. 2024.

\bibitem{hollmann2024large}
Hollmann N, M{\"u}ller S, Hutter F.
\newblock Large language models for automated data science: Introducing caafe for context-aware automated feature engineering.
\newblock Advances in Neural Information Processing Systems. 2024;36.

\bibitem{han2024large}
Han S, Yoon J, Arik SO, Pfister T.
\newblock Large Language Models Can Automatically Engineer Features for Few-Shot Tabular Learning.
\newblock In: Forty-first International Conference on Machine Learning; 2024. .

\bibitem{toufiq2023harnessing}
Toufiq M, Rinchai D, Bettacchioli E, Kabeer BSA, Khan T, Subba B, et~al.
\newblock Harnessing large language models (LLMs) for candidate gene prioritization and selection.
\newblock Journal of Translational Medicine. 2023;21(1):728.

\bibitem{wang2024geneagent}
Wang Z, Jin Q, Wei CH, Tian S, Lai PT, Zhu Q, et~al.
\newblock GeneAgent: Self-verification Language Agent for Gene Set Knowledge Discovery using Domain Databases.
\newblock arXiv preprint arXiv:240516205. 2024.

\bibitem{shringarpure2024large}
Shringarpure SS, Wang W, Karagounis S, Wang X, Reisetter AC, Auton A, et~al.
\newblock Large language models identify causal genes in complex trait GWAS.
\newblock medRxiv. 2024:2024-05.

\bibitem{jin2024genegpt}
Jin Q, Yang Y, Chen Q, Lu Z.
\newblock Genegpt: Augmenting large language models with domain tools for improved access to biomedical information.
\newblock Bioinformatics. 2024;40(2):btae075.

\bibitem{Hou2023.03.11.532238}
Hou W, Shang X, Ji Z.
\newblock Benchmarking large language models for genomic knowledge with GeneTuring.
\newblock bioRxiv. 2025.
\newblock Available from: \url{https://www.biorxiv.org/content/early/2025/01/05/2023.03.11.532238}.

\bibitem{li2024exploring}
Li D, Tan Z, Liu H.
\newblock Exploring Large Language Models for Feature Selection: A Data-centric Perspective.
\newblock arXiv preprint arXiv:240812025. 2024.

\bibitem{huang2022towards}
Huang J, Chang KCC.
\newblock Towards reasoning in large language models: A survey.
\newblock arXiv preprint arXiv:221210403. 2022.

\bibitem{tam2024let}
Tam ZR, Wu CK, Tsai YL, Lin CY, Lee Hy, Chen YN.
\newblock Let Me Speak Freely? A Study on the Impact of Format Restrictions on Performance of Large Language Models.
\newblock arXiv preprint arXiv:240802442. 2024.

\bibitem{brown2020language}
Brown TB.
\newblock Language models are few-shot learners.
\newblock arXiv preprint ArXiv:200514165. 2020.

\bibitem{agrawal2022large}
Agrawal M, Hegselmann S, Lang H, Kim Y, Sontag D.
\newblock Large language models are few-shot clinical information extractors.
\newblock arXiv preprint arXiv:220512689. 2022.

\bibitem{kandpal2023large}
Kandpal N, Deng H, Roberts A, Wallace E, Raffel C.
\newblock Large language models struggle to learn long-tail knowledge.
\newblock In: International Conference on Machine Learning. PMLR; 2023. p. 15696-707.

\bibitem{liu2024lost}
Liu NF, Lin K, Hewitt J, Paranjape A, Bevilacqua M, Petroni F, et~al.
\newblock Lost in the middle: How language models use long contexts.
\newblock Transactions of the Association for Computational Linguistics. 2024;12:157-73.

\bibitem{kortukov2024studying}
Kortukov E, Rubinstein A, Nguyen E, Oh SJ.
\newblock Studying Large Language Model Behaviors Under Realistic Knowledge Conflicts.
\newblock arXiv preprint arXiv:240416032. 2024.

\bibitem{lewis2020retrieval}
Lewis P, Perez E, Piktus A, Petroni F, Karpukhin V, Goyal N, et~al.
\newblock Retrieval-augmented generation for knowledge-intensive nlp tasks.
\newblock Advances in Neural Information Processing Systems. 2020;33:9459-74.

\bibitem{tong2024can}
Tong Y, Li D, Wang S, Wang Y, Teng F, Shang J.
\newblock Can LLMs Learn from Previous Mistakes? Investigating LLMs' Errors to Boost for Reasoning.
\newblock arXiv preprint arXiv:240320046. 2024.

\bibitem{fairley2020international}
Fairley S, Lowy-Gallego E, Perry E, Flicek P.
\newblock The International Genome Sample Resource (IGSR) collection of open human genomic variation resources.
\newblock Nucleic acids research. 2020;48(D1):D941-7.

\bibitem{luo2021machine}
Luo X, Li F, Xu W, Hong K, Yang T, Chen J, et~al.
\newblock Machine learning-based genetic diagnosis models for hereditary hearing loss by the GJB2, SLC26A4 and MT-RNR1 variants.
\newblock EBioMedicine. 2021;69.

\bibitem{Moustafa2023}
Moustafa A. Genetic Ancestry; 2023.
\newblock \url{https://github.com/ahmedmoustafa/genetic-ancestry} [Accessed: May 6 2024].

\bibitem{hollmann2022tabpfn}
Hollmann N, M{\"u}ller S, Eggensperger K, Hutter F.
\newblock Tabpfn: A transformer that solves small tabular classification problems in a second.
\newblock arXiv preprint arXiv:220701848. 2022.

\bibitem{hegselmann2023tabllm}
Hegselmann S, Buendia A, Lang H, Agrawal M, Jiang X, Sontag D.
\newblock Tabllm: Few-shot classification of tabular data with large language models.
\newblock In: International Conference on Artificial Intelligence and Statistics. PMLR; 2023. p. 5549-81.

\bibitem{luo2024large}
Luo X, Rechardt A, Sun G, Nejad KK, Y{\'a}{\~n}ez F, Yilmaz B, et~al.
\newblock Large language models surpass human experts in predicting neuroscience results.
\newblock Nature human behaviour. 2024:1-11.

\bibitem{li2024dalk}
Li D, Yang S, Tan Z, Baik JY, Yun S, Lee J, et~al.
\newblock DALK: Dynamic Co-Augmentation of LLMs and KG to answer Alzheimer's Disease Questions with Scientific Literature.
\newblock arXiv preprint arXiv:240504819. 2024.

\end{thebibliography}

\end{document}